%% file: main.tex
\documentclass[runningheads]{llncs}

\usepackage[mobile]{eccv}

\usepackage{eccvabbrv}

\usepackage{graphicx}
\usepackage{booktabs}
\usepackage{xcolor}

\newcommand{\titleurl}[1]{%
  {\textrm{\textcolor{orange}{\texttt{\url{#1}}}}}%
}

\usepackage[accsupp]{axessibility}  % Improves PDF readability for those with disabilities.

\usepackage{hyperref}

\usepackage{orcidlink}
\usepackage{wrapfig}

\begin{document}

\title{MultiGen: Level-Design for Editable Multiplayer Worlds in Diffusion Game Engines}
\subtitle{\titleurl{https://ryanpo.com/multigen}}
\titlerunning{MultiGen}

\author{Ryan Po\inst{1} \and
David Junhao Zhang\inst{2} \and
Amir Hertz\inst{2} \and \\
Gordon Wetzstein*\inst{1} \and 
Neal Wadhwa*\inst{2}  \and 
Nataniel Ruiz*\inst{2}
}

\authorrunning{Po et al.}

\institute{Stanford University \and Google}

\maketitle
\begingroup
\renewcommand\thefootnote{}\footnotetext{* Equal contributions.}
\addtocounter{footnote}{-1}
\endgroup

\begin{abstract}
    Video world models have shown immense promise for interactive simulation and entertainment, but current systems still struggle with two important aspects of interactivity: user control over the environment for reproducible, editable experiences, and shared inference where players hold influence over a common world. To address these limitations, we introduce an explicit external memory into the system, a persistent state operating independent of the model's context window, that is continually updated by user actions and queried throughout the generation roll-out. Unlike conventional diffusion game engines that operate as next-frame predictors, our approach decomposes generation into Memory, Observation, and Dynamics modules. This design gives users direct, editable control over environment structure via an editable memory representation, and it naturally extends to real-time multiplayer rollouts with coherent viewpoints and consistent cross-player interactions.
\end{abstract}

% =========================
% Replace the template body (from \section{Introduction} onward) with this outline.
% Keep everything above \begin{abstract} / \end{abstract} as-is.
% =========================

\input{sections/intro}

\input{sections/related_work}
\input{sections/method}
\input{sections/level_design}
\input{sections/multiplayer}

\section{Discussion and Conclusions}
\label{sec:discussion}

\begin{wraptable}{r}{0.42\linewidth}
\vspace{-42pt}
\centering
\small
\setlength{\tabcolsep}{4pt}
\renewcommand{\arraystretch}{1.05}
\begin{tabular}{l c c c}
\toprule
\addlinespace[1pt]
$L$ & SSIM $\uparrow$ & PSNR $\uparrow$ & LPIPS $\downarrow$ \\
\midrule
2  & 0.709 & 27.6 & 0.121 \\
4  & 0.775 & 29.5 & 0.097 \\
8  & 0.783 & 29.8 & 0.094 \\
16 & 0.782 & 29.8 & 0.093 \\
32 & 0.789 & 30.0 & 0.089 \\
\addlinespace[1pt]
\bottomrule
\end{tabular}
\vspace{-4pt}
\caption{Context frame ablation.}
\label{tab:ablation_context_wraptable}
\vspace{-20pt}
\end{wraptable}

\vspace{-0.5em}
\subsubsection{Ablation.}
We perform an ablation study over the number of conditioning frames, varying $L \in {2,4,8,16,32}$ while keeping all other settings fixed. As shown in~\cref{tab:ablation_context_wraptable}, increasing the context length consistently improves fidelity, reflected by higher SSIM/PSNR and lower LPIPS.
\vspace{-0.5em}
\subsubsection{Limitations.}
Our approach relies on the explicit state for long-horizon consistency. Consequently, scene properties not represented in the map $M$ (e.g., textures or small objects) are not explicitly preserved when revisiting the same region, which can lead to appearance inconsistencies. The dynamics model is also imperfect, so small pose errors may accumulate over long rollouts. However, actions remain aligned with plausible motion and the overall gameplay experience is preserved. Finally, visual appearance is bounded by the training distribution and may not generalize to styles far outside the collected VizDoom trajectories.
\vspace{-1.5em}
\subsubsection{Conclusion.}
We present a MultiGen, diffusion game engine built around explicit external memory, enabling level-conditioned gameplay generation and real-time multiplayer interaction in a shared world. Our system combines (i) external memory that stores map geometry and the evolving set of player poses, (ii) an observation model conditioned on ray-traced disparity and actions, and (iii) a lightweight dynamics model that updates pose to advance the roll-out. This decomposition supports controllable level design from coarse layouts and improves structural adherence, while also providing a scalable interface for multi-player roll-outs where each player generates a consistent first-person view conditioned on the same underlying state. We believe this modular, memory-centric formulation is a step toward more controllable and extensible generative game engines.

\section*{Acknowledgments.} 
We would like to thank Yang Zheng, Hadi Alzayer, Zizhang Li, Lior Yariv, Ilker Oguz, Seung-Woo Nam, Julian Quevedo, Bryan Chiang, Eric Li, and Alec Yiu for help with testing real-time demo and fruitful discussions. R.P. is supported by the Stanford Graduate Fellowship.

% Restate thesis: external memory is the enabling primitive; multiplayer + level design are two instances.

\bibliographystyle{splncs04}
\bibliography{main}
\end{document}

%% file: sections/intro.tex
\section{Introduction}
\label{sec:intro}
Recent advances in generative models have made real-time, action-controllable video generation increasingly practical, enabling interactive rollouts that resemble explorable ~\cite{hafner2020dreamcontrollearningbehaviors, hafner2022masteringataridiscreteworld, hafner2024masteringdiversedomainsworld, hafner2025trainingagentsinsidescalable, mao2025yume, genie3, parkerholder2024genie2, Bruce2024GenieGI}. However, most video world models are still fundamentally single-user experiences~\cite{Li2025HunyuanGameCraftHI, Yu2025GameFactoryCN, zhang2025matrix, Che2024GameGenXIO}: they generate the world on the fly with only implicit internal state, making it difficult to support \textbf{shared} worlds where multiple players can reliably interact through a common underlying state. In parallel, the same limitation restricts \textbf{control}: creators have little ability to specify the structure of an environment up front, and long rollouts can become hard to steer, hard to reproduce, and misaligned with user intent.

\begin{figure*}[t]
    \centering
    \includegraphics[width=\textwidth]{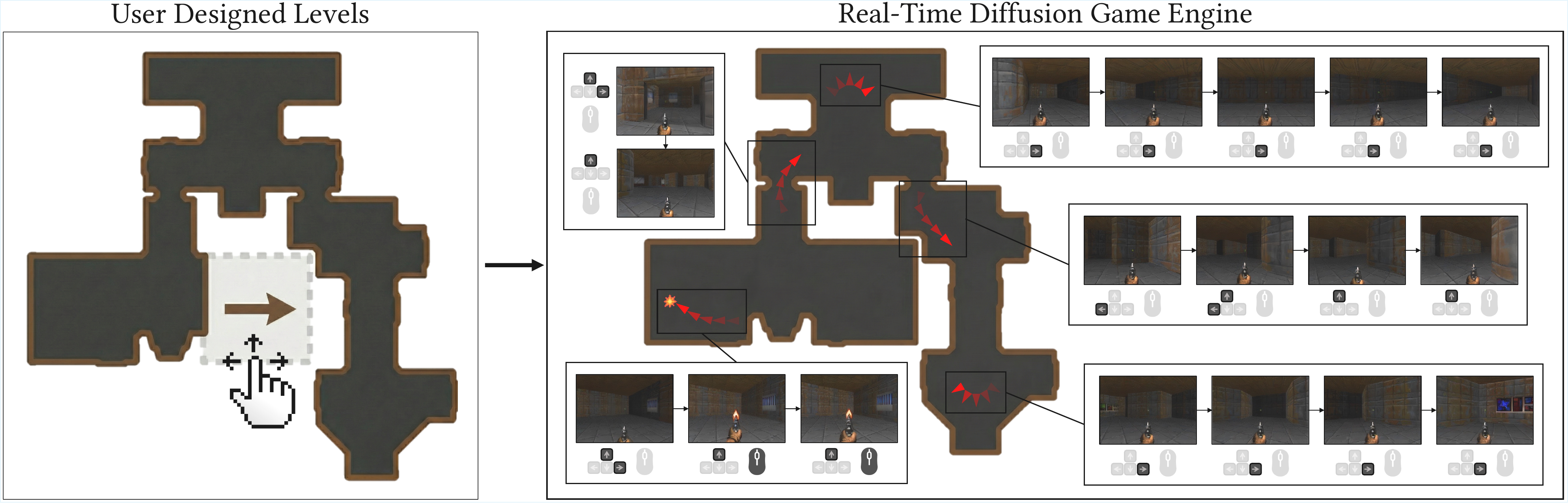}
    \caption{\textbf{Level Design via Editable Memory.} Users define a level through coarse 2D geometry (left). During inference, the diffusion model generates first-person observations consistent with the top-down level layout (right). }
    \label{fig:level_spread}
    \vspace{-2em}
\end{figure*}

We argue that both multiplayer interactivity and environment authoring point to a missing primitive: an explicit external memory that persists beyond the model’s context window and is updated by user actions. With external memory, multiple agents can read and write a shared state to support consistent cross-viewpoint interaction, and users can directly edit the state to control global structure. In this paper, we introduce a memory-augmented diffusion world model that leverages this idea for real-time interactive generation.

We use Doom~\cite{doom1993} as a controlled but expressive testbed. Doom offers rich first-person dynamics while retaining a clear notion of level layout, making it well suited for studying shared, long-horizon rollouts. To provide a simple and general interface for environment control, we represent world structure as a top-down 2D minimap that can be authored or edited before interaction begins. This minimap serves as an external memory blueprint that anchors generation while leaving the model free to synthesize detailed observations and moment-to-moment interactions.

Implementing external memory requires moving beyond the single-model paradigm used by prior diffusion game engines~\cite{Valevski2024DiffusionMA, Yu2025GameFactoryCN, zhang2025matrix, Che2024GameGenXIO, mao2025yume, genie3, parkerholder2024genie2, Bruce2024GenieGI}. Instead, we decompose the system into three specialized modules: \textbf{Memory}, \textbf{Observation}, and \textbf{Dynamics}. The memory module maintains a persistent state (including the minimap and agent states). The observation module generates the next visual observation conditioned on the memory readout and recent history. The dynamics module updates state given actions and observations. This separation makes long-horizon structure easier to maintain and, critically, enables multiplayer in a natural way: multiple agents act on the same shared memory, and the model can render coherent observations from one or more viewpoints with interaction effects between players.

Together, these choices support (i) \textbf{real-time multiplayer rollouts} grounded in shared external memory, and (ii) a simple \textbf{level-design workflow} where users specify or edit a minimap and obtain consistent, reproducible interactive sessions. In summary, our contributions are:
\begin{itemize}
    \item We introduce an external-memory-based formulation for diffusion world models that supports shared state updates from user actions, enabling consistent long-horizon interactive rollouts.
    \item We propose a modular architecture with Memory, Observation, and Dynamics modules, replacing the single-model paradigm and providing a clean interface for read/write external memory.
    \item We demonstrate two applications enabled by external memory: editable environment design via a minimap blueprint, and real-time multiplayer interaction with coherent cross-viewpoint behavior, and evaluate each against relevant baselines.
\end{itemize}

%% file: sections/related_work.tex
\section{Related Work}
\label{sec:related}

\subsubsection{Video Diffusion Models.} 
Current state-of-the-art in video generation models have mostly been set by large-scale bidirectional diffusion transformers~\cite{Peebles2022ScalableDM, videoworldsimulators2024}, demonstrating remarkable capabilities in synthesizing high-quality, complex video clips. The core mechanism relies on full spatiotemporal attention applied to all tokens~\cite{videoworldsimulators2024}, as every video token is denoised simultaneously~\cite{Blattmann2023StableVD, Blattmann2023AlignYL, videoworldsimulators2024, Li2024AutoregressiveIG, Gupta2023PhotorealisticVG, HaCohen2024LTXVideoRV, Ho2022ImagenVH, Ho2022VideoDM, Kong2024HunyuanVideoAS, Polyak2024MovieGA, Villegas2022PhenakiVL, Wang2025WanOA, Hong2022CogVideoLP, Yang2024CogVideoXTD, BarTal2024LumiereAS, Cai2025MixtureOC,zhang2025show}, generating videos of fixed lengths. However, these models are inherently non-causal and constrained to fixed-length generation.

\subsubsection{Autoregressive Video Models.} Autoregressive approaches in video generation factorizes the joint distribution over all frames into, $p(x^{1:N}) = \prod^N_{i=1} p(x^i | x^{<i})$. This formulation naturally aligns with the causality of time, as video frames are generated sequentially, making autoregressive models well suited for interactive simulation~\cite{Shin2025MotionStreamRV, Huang2025SelfFB}. Conventional autoregressive video models rely on direct next-token prediction of discrete video tokens~\cite{Bruce2024GenieGI, Kondratyuk2023VideoPoetAL, Ren2025NextBP, Wang2024LoongGM, Weissenborn2019ScalingAV, Yan2021VideoGPTVG}. However, the performance of discretized AR models often lag behind diffusion models. To address this, recent works have explored training strategies combining autoregression and diffusion~\cite{Gao2024Ca2VDMEA, Gu2025LongContextAV, Guo2025LongCT, Hu2024ACDiTIA, Jin2024PyramidalFM, Li2024ARLONBD, Liu2024MarDiniMA, Liu2024RedefiningTM, Weng2023ARTVAT, Zhang2025TestTimeTD, Zhang2025GenerativePA}. Some works train conditional diffusion models that denoise next frames condition on past clean frames~\cite{Zhang2025FrameCP, Huang2025SelfFB}, while other approaches introduce per-frame independent noise levels during training~\cite{Chen2024DiffusionFN, Song2025HistoryGuidedVD, Valevski2024DiffusionMA}, allowing for AR inference. 

\subsubsection{Diffusion Game Engines.}
Our work is most closely related to recent efforts that repurpose diffusion models as real-time game engines or world simulator~\cite{Li2025HunyuanGameCraftHI, Yu2025GameFactoryCN, zhang2025matrix, Che2024GameGenXIO, mao2025yume, genie3, parkerholder2024genie2, Bruce2024GenieGI}. GameNGen~\cite{Valevski2024DiffusionMA} generates gameplay by conditioning an image diffusion model on a short history of previous frames and the next action, using diffusion as an autoregressive next-frame generator. We build on this by introducing a modular diffusion game engine with explicit external memory. This design provides a persistent reference to the authored level layout, improving structural adherence over long horizons and enabling practical level-conditioned generation from coarse user edits.

\subsubsection{Video Models with External Memory.} While existing video frameworks typically construct memory using explicit 3D representations \cite{Wu2025VideoWM, Deng2024StreetscapesLC, zhao2025spatia} or manage input conditions through compressed context windows and KV caching \cite{huang2025memoryforcing,wu2026infiniteworld,Gu2025LongContextAV,yu2025contextmemory,li2025vmem, hong2025relic, wu2025pack}, our approach develops an external memory that persists beyond the model’s context window. Updated by user actions, this memory provides a shared state for multiple agents to read and write, ensuring consistent cross-viewpoint interaction while allowing users to directly edit the state to govern the global structure.

\subsubsection{Game Generation.} Our work is also related to the field of Game Generation as a whole. A wide range of approaches have been proposed for directly generating game content ~\cite{Summerville2017ProceduralCG}. Recent works have utilized Generative Adversarial Networks (GANs) to great effect for generating game levels and interactiive environments~\cite{Volz2018EvolvingML, Kumaran2020GeneratingGL, Schubert2021TOADGANAF, Kim2020LearningTS}. While earlier works explore a wide range of methods for generating game content~\cite{Treanor2015AIbasedGD, Snodgrass2014ExperimentsIM, Guzdial2021GameLG, Summerville2016SuperMA}. Other works also explore the use of diffusion models for game generation~\cite{Valevski2024DiffusionMA, Zhou2024TheET} and LLMs for designing game mechanics~\cite{Sudhakaran2023MarioGPTOT, Todd2023LevelGT, Nasir2023PracticalPT, Hu2024GameGV, Zala2024EnvGenGA, Anjum2024TheIS, Chung2024PatchviewLW, Chung2024ToytellerTW}. These approaches primarily focus on generating the game content itself such as world layouts, assets, or mechanics. Our model instead operates like a game engine and gives users additional control over the environment while generating every observed frame online through a diffusion model.

%% file: sections/method.tex
\section{Method}
\label{sec:method}

We model an interactive game rollout as a sequence of actions and observations over discrete timesteps $t \in \{0,1,\dots,T\}$. Let $a_t \in \mathcal{A}$ denote the agent action at timestep $t$, and let $o_t \in \mathbb{R}^{H \times W \times C}$ denote the rendered observation (image frame) at timestep $t$. A rollout is therefore represented as an alternating sequence
\begin{equation}
    \tau \;=\; \Big( o_0, a_0, o_1, a_1, \dots, a_{T-1}, o_T \Big).
\end{equation}
Given a history of past observations and actions, the goal of a diffusion game engine is to predict the distribution of the next observation,
\begin{equation}
    p\!\left(o_{t+1} \mid o_{\le t}, a_{\le t}\right),
\end{equation}
and to enable closed-loop simulation by repeatedly sampling $o_{t+1}$ and feeding it back as input at the next step.

Following GameNGen~\cite{Valevski2024DiffusionMA}, we instantiate this predictor as a conditional generative model that maps a representation of the current ``state'' together with the next action to a distribution over the next frame. In the original GameNGen~\cite{Valevski2024DiffusionMA} formulation, the state is implicit and is represented solely by a fixed-length window of the most recent observed frames. Concretely, letting
\begin{equation}
    S_t \;=\; o_{t-L+1:t} \;=\; \big(o_{t-L+1}, \dots, o_t\big)
\end{equation}
denote an $L$-frame context, the next observation is generated by querying a diffusion model conditioned on the current state $S_t$ and action $a_t$:
\begin{equation}
    o_{t+1} \sim p_\theta\!\left(o \mid S_t, a_t\right).
\end{equation}
Intuitively, $S_t$ provides the visual history needed to maintain appearance and short-term temporal continuity, while $a_t$ specifies the agent control signal that drives the transition.

\begin{figure}[t]
    \centering
    \includegraphics[width=0.85\textwidth]{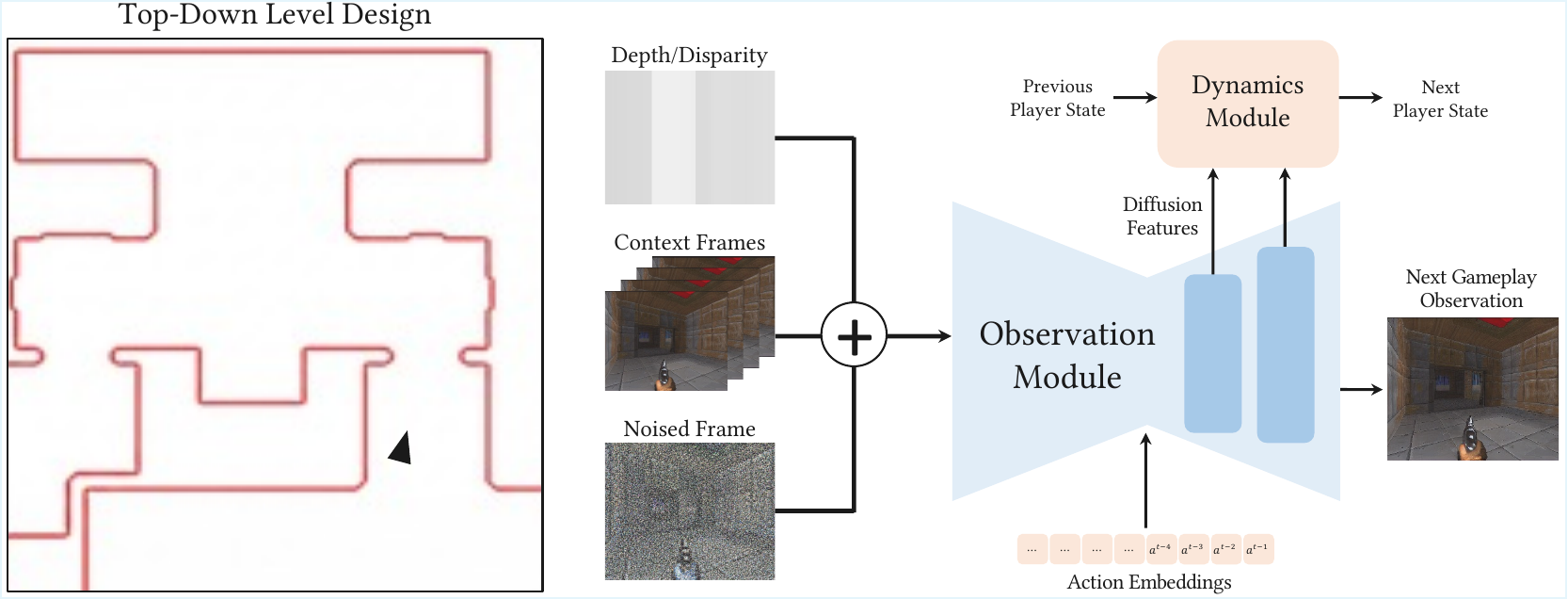}
    \caption{\textbf{Method overview.} We introduce an explicit external memory and factor the diffusion game engine into three modules: \textbf{\textit{Memory}} (map geometry and pose; \cref{sec:memory}), \textbf{\textit{Observation}} (next-frame generation conditioned on history and memory readouts; \cref{sec:obs}), and \textbf{\textit{Dynamics}} (pose update for state progression; \cref{sec:dynamics}).}
    \label{fig:method}
\end{figure}

In our work, we retain this interactive generative modeling setup but go beyond the conventional ``frames-as-state'' design~\cite{hafner2020dreamcontrollearningbehaviors, hafner2022masteringataridiscreteworld, hafner2024masteringdiversedomainsworld, hafner2025trainingagentsinsidescalable, mao2025yume, genie3, parkerholder2024genie2, Bruce2024GenieGI} by introducing an explicitly controlled external memory~\cite{Wu2025VideoWM, Deng2024StreetscapesLC, zhao2025spatia}. Rather than requiring a single network to both (i) maintain long-horizon information about the environment and agent configuration and (ii) generate high-dimensional observations, we factor the diffusion game engine into three components: a \textbf{\textit{memory module}} that stores structured map geometry and the player pose (\ref{sec:memory}), an \textbf{\textit{observation module}} that predicts the next frame conditioned on the visual context and memory-derived geometric signals (\ref{sec:obs}), and a \textbf{\textit{dynamics module}} that updates the player pose to advance the state over time (\ref{sec:dynamics}). This modularization separates persistent, low-dimensional game state from high-dimensional image generation, leading to a more controllable and interpretable simulation process.

\subsection{Memory Module}
\label{sec:memory}

While GameNGen~\cite{Valevski2024DiffusionMA} represents the simulator state implicitly as a window of recent frames, we instead maintain an explicit state that separates persistent, low-dimensional game information from high-dimensional observations. Concretely, at timestep $t$ we define the state as
\begin{equation}
    S_t = \big(M,\; p_t,\; o_{t-L+1:t}\big),
    \label{eq:state_def}
\end{equation}
where $o_{t-L+1:t} = (o_{t-L+1},\dots,o_t)$ denotes an $L$-frame visual context, $M$ denotes the (static) level map, and $p_t$ denotes the player pose.

We represent the map $M$ as a set of 2D vertices and line segments defining the walkable layout and walls:
\begin{equation}
\begin{aligned}
    M &= (V, E), \\
    V &= \{v_i\}_{i=1}^{N_v}, \qquad v_i \in \mathbb{R}^2, \\
    E &= \{e_j\}_{j=1}^{N_e}, \qquad e_j \in \{1,\dots,N_v\}^2,
\end{aligned}
\label{eq:map_def}
\end{equation}
where each $e_j=(u_j,w_j)$ indexes an (undirected) line segment connecting vertices $v_{u_j}$ and $v_{w_j}$.

Intuitively, $M$ serves as a reliable, persistent external reference for the generative game engine. In a ``frames-as-state'' design, all information about the environment must be preserved implicitly within a finite visual context $o_{t-L+1:t}$; as rollouts grow longer, relevant layout cues may fall out of the buffer, forcing the model to hallucinate or re-infer structure from incomplete evidence. In contrast, the map $M$ is time-invariant and can always be consulted to provide a coarse but stable description of the current level. This persistent signal simplifies long-horizon consistency by giving the model an explicit representation of global geometry that does not degrade with context length.

Player information is parameterized by their euclidean coordinates and yaw angle,
\begin{equation}
    p_t = (x_t, y_t, \theta_t) \in \mathbb{R}^2 \times \mathbb{S}^1,
    \label{eq:pose_def}
\end{equation}
with $(x_t,y_t)$ denoting the player location in map coordinates and $\theta_t$ denoting the facing direction.

The memory module maintains the static map $M$ and the evolving pose $p_t$ throughout the rollout, and provides these quantities to the observation and dynamics modules as part of $S_t$. In the following sections, we describe how the map and pose are used to compute auxiliary geometric signals (e.g., a ray-traced 1D depth observation) for frame prediction, and how player information is updated over time.

% \begin{figure}
%     \centering
%     \includegraphics[width=\linewidth]{Screen Shot 2026-01-03 at 1.16.02 AM.png}
%     \caption{Memory stores map geometry and player information, providing ray-traced disparity for conditioning. The observation UNet generates the next frame given context and action, and the dynamics transformer predicts the pose update to advance the rollout.}
%     \label{fig:method}
% \end{figure}
\subsection{Observation Module}
\label{sec:obs}

The observation module generates the next visual observation conditioned on recent context, a geometric readout from external memory, and the next action. At timestep $t$, we model
\begin{equation}
    o_{t+1} \sim p_\phi\!\left(o \mid o_{t-L+1:t},\, r_t,\, a_t \right),
    \label{eq:obs_conditional}
\end{equation}
where $r_t$ is a geometric conditioning signal derived from the memory module (\cref{sec:memory}).

\subsubsection{Geometric conditioning via disparity.}
Given the current pose and map, the memory module ray-traces a 1D depth vector within the agent's field of view and converts it to disparity (inverse depth) to emphasize near-field structure. We then feed geometry to the UNet by mapping the 1D disparity~\cite{Deng2024StreetscapesLC} to a spatial tensor at the UNet input resolution and concatenating it as additional channels alongside the $L$ context frames.

\subsubsection{Action conditioning.}
We represent the discrete action $a_t$ with a learned embedding and inject it into the observation UNet through cross-attention conditioning tokens. This allows the denoiser to modulate generation according to the agent's control input without changing the convolutional input interface.

\subsubsection{Diffusion objective.}
We instantiate $p_\phi$ as a diffusion model~\cite{ho2020denoisingdiffusionprobabilisticmodels} trained with the standard velocity-parameterization objective~\cite{salimans2022progressivedistillationfastsampling}. Given a ground-truth next frame $o_{t+1}$, we sample a diffusion timestep $\tau$, construct a noised version according to the forward process, and train the UNet to match the corresponding velocity target. Concretely, we optimize
\begin{equation}
    \mathcal{L}_{\mathrm{obs}}
    = \mathbb{E}_{t,\tau}\left[
        \left\| v_\phi(\cdot \mid o_{t-L+1:t}, r_t, a_t;\tau) - v^\star(o_{t+1}, \tau) \right\|_2^2
    \right],
    \label{eq:obs_loss}
\end{equation}
where $v^\star$ denotes the standard velocity target associated with the chosen diffusion parameterization.

\subsubsection{Noised-context training for drift robustness.}
During training, the observation module conditions on ground-truth context frames, whereas at test time it conditions on its own generated history. To reduce this train--test mismatch~\cite{Huang2025SelfFB,Cui2025SelfForcingTM,Liu2025RollingFA,po2025baggerbackwardsaggregationmitigating}, we follow prior work~\cite{Chen2024DiffusionFN, Valevski2024DiffusionMA} and corrupt all context frames with Gaussian noise at a randomly sampled noise scale during training. This exposes the model to imperfect histories and improves robustness under long autoregressive rollouts.

\subsection{Dynamics Module}
\label{sec:dynamics}

To advance an interactive rollout beyond single-step prediction, the system must update the agent state used by external memory. In our setting, this reduces to predicting the next player pose
$p_t = (x_t, y_t, \theta_t)$, where $\theta_t$ is wrapped to a canonical range.

\subsubsection{Inputs.}
The dynamics module consumes (i) the action $a_t$, (ii) the same geometric conditioning signal $r_t$ used by the observation module, and (iii) intermediate UNet features produced while denoising the next frame (e.g., a bottleneck feature map). We aggregate these UNet features into a fixed-dimensional representation (e.g., via pooling) and concatenate them with embeddings of pose, action, and geometry.

\subsubsection{Lightweight transformer dynamics.}
We implement dynamics as a small transformer encoder that predicts an incremental pose update:
\begin{equation}
    \Delta \hat{p}_t
    = \mathcal{D}_\psi\!\left(p_t,\, a_t,\, r_t,\, f_t\right),
    \label{eq:dynamics_model}
\end{equation}
where $f_t$ denotes the aggregated UNet feature representation. We apply this increment to obtain $\hat{p}_{t+1}$, using angle wrapping for the orientation component.

\subsubsection{Training objective and state update.}
We supervise dynamics using ground-truth poses from the environment. In practice, we use an $\ell_2$ loss on translation and a wrapped-angle error on orientation. After predicting $\hat{p}_{t+1}$, the external memory state is advanced by updating the pose and shifting the visual context window to include the newly generated frame.

\subsection{Inference}
\label{sec:inference}

At inference time, the system functions as an interactive simulator that repeatedly maps the current state and an action to the next observation and updated state:
\begin{equation}
    (S_t, a_t) \mapsto (\hat{o}_{t+1}, S_{t+1}),
    \label{eq:inference_interface}
\end{equation}
where $S_t = (M, p_t, o_{t-L+1:t})$ contains the static map $M$, the current pose $p_t$, and an $L$-frame context window.

For each timestep, we (1) query external memory to obtain the geometric readout $r_t$ from the current pose and map, (2) sample the next frame $\hat{o}_{t+1}$ using the diffusion observation model conditioned on the context, geometry, and action, and (3) update the pose with the dynamics module using the action, geometry, and intermediate UNet features. To stabilize long rollouts, we use history guidance~\cite{Song2025HistoryGuidedVD}: the conditional branch receives the clean context frames, while the unconditional branch receives a noised version of the context, encouraging fidelity to recent history while retaining robustness to imperfect inputs. The state is then advanced by shifting the context window and updating the pose for the next step.

%% file: sections/level_design.tex
\begin{figure*}[t]
    \centering
    \includegraphics[width=\linewidth]{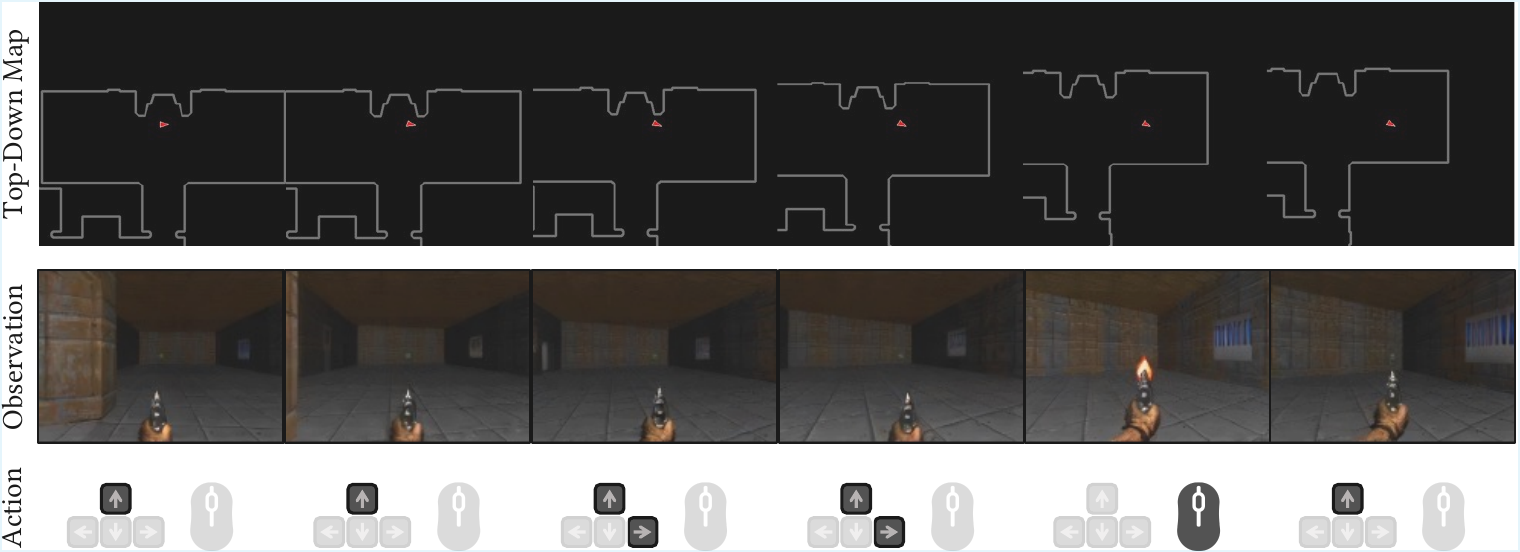}
    \vspace{-1.0em}
    \caption{Example rollouts under an authored map and action sequence. Top: minimap $M$ with pose $p_t$ (red arrow). Middle: generated first-person observations $\hat{o}_t$. Bottom: actions $a_t$. The viewpoint evolves coherently with the action inputs while adhering with the designed layout.}
    \label{fig:level_results}
    \vspace{-1.5em}
\end{figure*}

\section{Application I: Level Design}
\label{sec:level-design}

A primary advantage of an external memory is that it provides a direct handle for modifying the underlying structure of the world. By defining the world explicitly with course map structures, users can directly influence the structure of the environment before inference even begins. We evaluate this capability on \textbf{\textit{level-conditioned gameplay generation}}: synthesizing interactive first-person rollouts that remain consistent with a user-authored level layout. In this task, the model is given (i) a top-down map $M$ specified as coarse line geometry (vertices and wall segments), (ii) an initial player pose $p_0$, and (iii) a sequence of actions $\{a_t\}_{t=0}^{T-1}$, and generates a corresponding sequence of observations $\{\hat{o}_t\}_{t=1}^{T}$. Unlike implicit-state diffusion game engines that must infer global structure from a finite history, our approach can query $M$ at every timestep through an external-memory readout, providing a persistent geometric reference throughout the roll-out.

\subsection{Level Design Dataset}
\label{sec:level-data}
To train our model to generalize across a diverse set of level layouts, we generate gameplay sequences on 100 procedurally generated maps with randomized structure. We create these maps using the Obsidian map generator~\cite{ObsidianLevelGeneratorWebsite}, which produces varied layouts while preserving valid Doom geometry. We then deploy a pre-trained Doom agent~\cite{lample2018playingfpsgamesdeep} to explore the resulting maps, collecting over 10 million gameplay frames paired with the corresponding actions and player poses. This dataset exposes the model to a wide range of corridor and room configurations, encouraging it to rely on external memory for global structure rather than memorizing a small set of fixed levels.

\begin{table}[t]
\centering
\setlength{\tabcolsep}{3.5pt}
\caption{SSIM/PSNR/LPIPS between generated frames and ground truth. Our method consistently outperforms baselines, with the largest gains appearing in later rollout stages where consistent memory is the most important.}
\vspace{-1em}
\label{tab:ssim_psnr_lpips}
{\fontsize{8.5pt}{10pt}\selectfont
\begin{tabular}{l c c c c c c c c c}
\toprule
& \multicolumn{3}{c}{SSIM $\uparrow$} & \multicolumn{3}{c}{PSNR $\uparrow$} & \multicolumn{3}{c}{LPIPS $\downarrow$} \\
\cmidrule(lr){2-4} \cmidrule(lr){5-7} \cmidrule(lr){8-10}
Method & All & 1--128 & 128--256 & All & 1--128 & 128--256 & All & 1--128 & 128--256 \\
\midrule

IP-Adapter~\cite{ye2023ipadaptertextcompatibleimage} & 0.415 & \best{0.433} & 0.396 & 18.74 & \best{20.30} & 17.19 & 0.488 & 0.397 & 0.578 \\
ControlNet~\cite{Zhang2023AddingCC} & 0.411 & 0.422 & 0.401 & 18.51 & 19.58 & 17.45 & 0.524 & 0.453 & 0.596 \\
GameNGen~\cite{Valevski2024DiffusionMA}   & 0.405 & 0.427 & 0.384 & 18.77 & 20.23 & 17.33 & 0.471 & \best{0.379} & 0.562 \\
Ours (MultiGen)      & \best{0.418} & 0.429 & \best{0.406} & \best{19.32} & 20.06 & \best{18.59} & \best{0.453} & 0.400 & \best{0.505} \\
\bottomrule
\end{tabular}
}
\vspace{-1.5em}

\end{table}

\begin{figure}
    \centering
    \includegraphics[width=\textwidth]{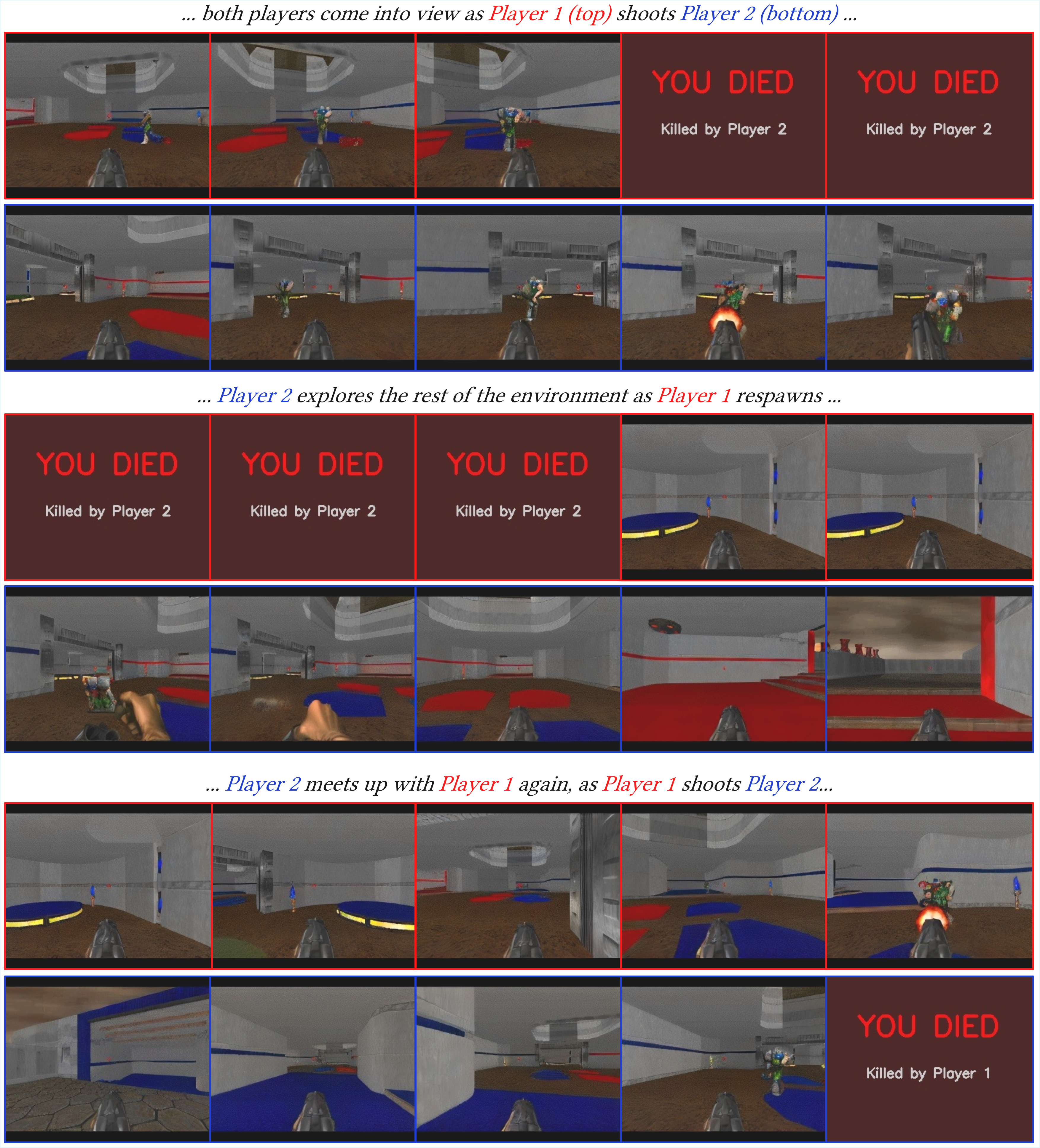}
    \caption{\textbf{Example Two-Player Gameplay Roll-out.} Our method generates consistent first-person views for both players by maintaining a shared world memory. The roll-out shows a short two-player interaction: the players meet, and Player 1 kills Player 2, after which Player 2 is removed from the shared state. Player 1 then explores the map while Player 2 respawns and is re-added to the shared state. The players meet again, and Player 1 kills Player 2 once more. Note that both views are are consistent with each other, as actions from one player directly effects the next-frame observation generated by the other model. All game play frames are generated using the observation module. Frames shown during player death are not part of the model output, and are only shown for illustrative purposes.}

    \label{fig:two_player_respawn}
\end{figure}

\subsection{Results}
\label{sec:level-qual}

\cref{fig:level_spread} shows representative rollouts conditioned on a designed map and action sequence. The top row visualizes the authored layout and the evolving player pose, the middle row shows generated first-person observations, and the bottom row shows the applied action inputs. Across these examples, the viewpoint evolves consistently with the action inputs, while adhering to the general structure of the layout specified by $M$. Qualitatively, this demonstrates that coarse user geometry is sufficient to anchor long rollouts: the model maintains stable corridor structure, respects turns represented by the map, and avoids any structural drift that occurs when global layout must be inferred from a limited frame-based visual history.

\subsection{Evaluations}
\label{sec:level-quant}

We compare against relevant baselines. First, we include GameNGen~\cite{Valevski2024DiffusionMA}, which models the entire engine as a single diffusion network without external memory, representing environment state implicitly through a finite window of past frames. We also compare against alternative approaches for conditioning on external state, specifically ControlNet~\cite{zhang2023addingconditionalcontroltexttoimage}  and IP-Adapter~\cite{ye2023ipadaptertextcompatibleimage}, which condition on the top-down minimap. For all methods, we initialize from the same initial observation and pose and roll out under the same action sequence for $T$ steps. We then measure similarity to the ground-truth observations from the underlying simulator using SSIM (structural similarity) and LPIPS (perceptual distance), reporting averages over early and late rollout ranges to assess long-horizon stability.

As shown in \cref{tab:ssim_psnr_lpips}, conditioning on external memory improves structural consistency, with larger gains in later rollout segments where implicit-state baselines are more prone to drift. Access to the authored map provides a stable geometric reference that anchors generation to the true layout over time, whereas the memory-free baseline increasingly hallucinates layout changes that compound under autoregressive sampling.

%% file: sections/multiplayer.tex
\section{Application II: Multiplayer Interaction}
\label{sec:multiplayer}

The external memory not only enables editability of the environment, it also naturally extends to act as a shared state that multiple agents can condition on and update in during generated roll-outs. We evaluate this capability on \textbf{\textit{multiplayer gameplay generation}}: synthesizing synchronized first-person rollouts for multiple players interacting within the same environment. In this application, the model is given (i) a shared top-down map $M$, (ii) initial player poses $\{p_0^{(i)}\}_{i=1}^{N}$, and (iii) per-player action sequences $\{a_t^{(i)}\}_{t=0}^{T-1}$, and generates per-player observations $\{\hat{o}_t^{(i)}\}_{t=1}^{T}$. During inference, all players query and modify the same external memory state, so that one player’s actions can influence what other players observe. This is in direct contrast with ``frames-as-state'' models~\cite{hafner2020dreamcontrollearningbehaviors, hafner2022masteringataridiscreteworld, hafner2024masteringdiversedomainsworld, hafner2025trainingagentsinsidescalable, mao2025yume, genie3, parkerholder2024genie2, Bruce2024GenieGI, enigma2025multiverse}, where state is entangled with the local observation history of every player, making cross-player consistency difficult to maintain over long horizons.

\begin{figure}[t]
    \centering
    \includegraphics[width=\textwidth]{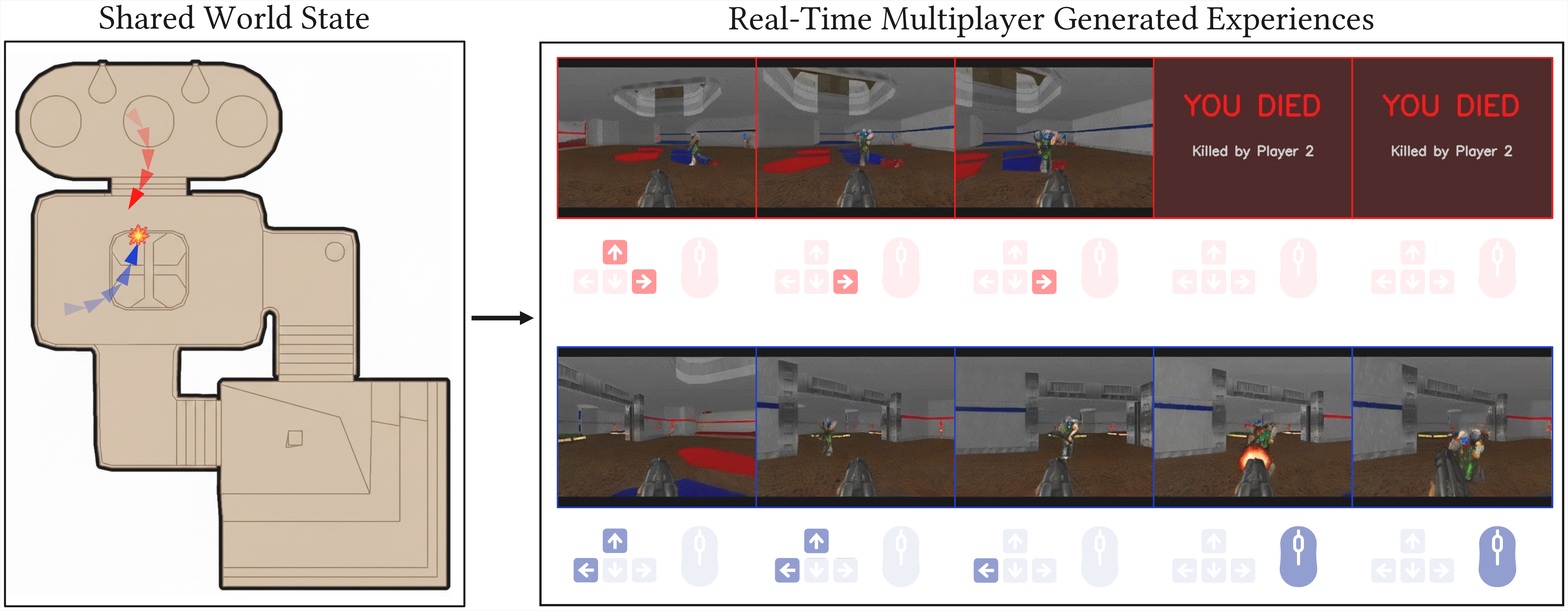}
    \caption{\textbf{Real-Time Interactive Multiplayer Generative Experiences.} A shared consistent world state (left) enables consistent multiplayer generative experiences (right). Our method leverages a diffusion model conditioned on past frame observations, the next player action, and the external world state to generate gameplay roll-outs in real-time. The shared world state enables meaningful interactions between players, such as one player killing another (right).}
    \label{fig:method}
    \vspace{-1.5em}
\end{figure}

\subsection{Shared Memory for Multiplayer Roll-outs}
\label{sec:mp-memory}

In our multiplayer setting, the shared world state is represented explicitly by the external memory: the static map layout $M$ together with the set of active player poses $\{p_t^{(i)}\}_{i=1}^{N_t}$. Generation is \textbf{distributed}: each player runs their own copy of the Observation and Dynamics modules, while all players read from and write to the same shared memory. At each timestep, every player submits an action $a_t^{(i)}$ and queries the shared state to obtain viewpoint-specific conditioning signals, including geometric depth/disparity from the map and information about other players that are visible from that pose. Conditioned on these shared-memory readouts, each player's Observation module generates the next first-person frame $\hat{o}_{t+1}^{(i)}$ for that player. After all players have generated their next frames, we update the shared state by applying each player's action through their Dynamics module, advancing the set of poses (and updating which players are active).

Crucially, this distributed design supports an arbitrary number of players without changing the model interface. In contrast, a simple joint baseline that models all viewpointsin a single ``split-screen'' observation~\cite{enigma2025multiverse} must fix the number of players at training time. Our approach naturally supports gameplay events such as player death and respawn by removing or reintroducing player poses in the shared memory.

\subsection{Multiplayer Dataset}
\label{sec:mp-dataset}
Because each model instance only generates its own first-person viewpoint, we can train on standard single-view gameplay data: consistent multiplayer viewpoints arise from shared external memory, without requiring explicit multi-view supervision. We build on ViZDoom~\cite{Wydmuch2019ViZdoom} and collect simulated Doom deathmatch sequences in which one pre-trained agent~\cite{lample2018playingfpsgamesdeep} plays against four identical agents. In total, we gather over 10 million gameplay frames. Each sequence records the ego-agent's action, the poses of all active players, and the map layout $M$. For this multiplayer study, we train and evaluate on a single map.

\subsection{Results}
\label{sec:mp-results}
\cref{fig:two_player_respawn} shows a two-player rollout with a typical gameplay loop: the players start at different map locations, approach until they come into each other's view, and Player~1 kills Player~2, removing Player~2 from the shared state and triggering a death animation from Player~2's viewpoint while Player~1 continues moving. After a delay, Player~2 respawns (re-entering the shared state), the players meet again, and Player~2 is killed in a second encounter. Beyond visual fidelity, the striking result is that these interactions remain \textbf{mutually consistent} across viewpoints despite being generated autoregressively: when a player should be visible, they appear with the correct pose and location, and when they are dead or out of view, they disappear from the other player's camera while the victim still observes a coherent death/respawn sequence.

\begin{figure}[t]
    \centering
    \includegraphics[width=\textwidth]{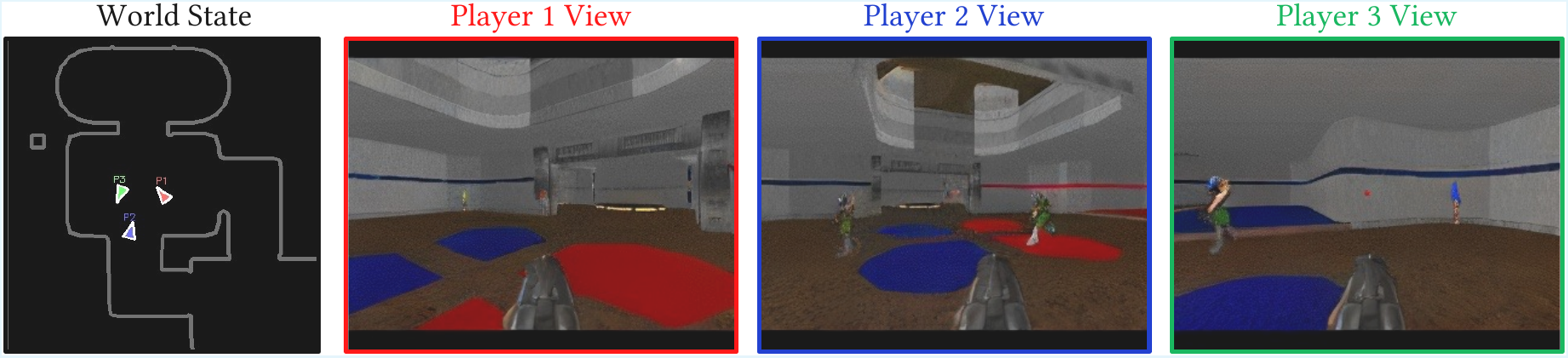}
    \caption{\textbf{Consistent Views in Three-Player.} Our distributed model system supports an arbitrary number of players in the shared state despite only being trained on data from a single viewpoint. Player positions are accurately represented in all three viewpoints in the example above.}

    \label{fig:three_player}
    \vspace{-1.5em}
\end{figure}

As described above, our method supports an arbitrary number of players by running separate Observation/Dynamics instances that share a common external world state. Importantly, each player's model runs independently, so adding more players does not slow down inference---we simply run additional per-player instances that read/write the same shared state. In our implementation, the full system runs at approximately \textbf{20 FPS} using a single NVIDIA A100 per player, making the resulting multiplayer rollouts not only consistent, but also interactive in real time. \cref{fig:three_player} shows consistent generations from three viewpoints, each aligned with the shared state: Player~1 sees no other players, Player~2 sees both Players~1 and~3, and Player~3 sees only Player~2.

\subsection{Evaluations}
\label{sec:mp-eval}

For quantitative comparisons, we compare to CotrolNet~\cite{zhang2023addingconditionalcontroltexttoimage} and IP-Adapter~\cite{ye2023ipadaptertextcompatibleimage}. Another natural baseline for multiplayer is to jointly model all player views with a single network, resembling a ``split-screen'' game~\cite{enigma2025multiverse}. To highlight the role of external memory, we compare our approach against a two-player split-screen diffusion model trained to predict both viewpoints on the same map without any explicit memory mechanism. In this baseline, multiplayer consistency must be learned purely from the limited observation histories of both players, since there is no shared structured state that persists beyond the context window.

We generate a set of random multiplayer gameplay trajectories from the environment and use them as ground truth. For each trajectory, we roll out all models by forcing each player to follow the ground-truth action stream, producing paired generated views for Player~1 and Player~2 over $T$ steps. We then measure multiplayer consistency via \textbf{\textit{opponent presence accuracy}}: whether the generated frame contains the other player when they should be visible. We use a pre-trained vision-language model (VLM) as an automated judge~\cite{openai2024openaio1card}. Concretely, for each timestep and each player's viewpoint, we query the VLM to determine whether the opponent is present in the generated frame, and compute accuracy by agreement with the same VLM judgement on the corresponding ground-truth frame. We report \textbf{accuracy}, \textbf{precision}, and \textbf{recall} (opponent visible = positive). Precision penalizes hallucinated opponents (false positives), while recall penalizes missed opponents when they should be visible (false negatives).

\begin{table}[t]
\centering
\setlength{\tabcolsep}{7pt}
\caption{Multiplayer consistency evaluation using opponent-presence detection. We report accuracy, precision, and recall of a VLM-based judge on generated frames against ground-truth visibility labels. MultiGen consistently outperforms all baselines.}
\label{tab:mp_acc_prec_rec}
\begin{tabular}{l c c c}
\toprule
Method & Accuracy $\uparrow$ & Precision $\uparrow$ & Recall $\uparrow$ \\
\midrule
IP-Adapter~\cite{ye2023ipadaptertextcompatibleimage} &  62.12\% & 84.62\% & 40.73\% \\
ControlNet~\cite{zhang2023addingconditionalcontroltexttoimage} &  60.71\% & 82.16\% & 39.20\% \\
Split-screen~\cite{enigma2025multiverse} &  65.31\% & 86.86\% & 44.59\% \\
Ours (MultiGen)   & \best{75.38\%} & \best{88.12\%} & \best{65.07\%} \\
\bottomrule
\vspace{-2.5em}
\end{tabular}
\end{table}